\documentclass{article}

\usepackage{arxiv}
\usepackage{cite}
\usepackage{dcolumn}
\usepackage[utf8]{inputenc} 
\usepackage[T1]{fontenc}    
\usepackage{url}            
\usepackage{booktabs}       
\usepackage{amsfonts}       
\usepackage{nicefrac}       
\usepackage{microtype}      
\usepackage{lipsum}
\usepackage{graphicx}
\graphicspath{ {./images/} }
\usepackage{multirow}
\usepackage{flushend}
\usepackage[bookmarksnumbered=true]{hyperref} 
\hypersetup{
    colorlinks = true,
    linkcolor = blue,
    anchorcolor = blue,
    citecolor = blue,
    filecolor = blue,
    urlcolor = blue
    }

\title{Classifier Combination Approach for Question Classification for Bengali Question Answering System}

\author{
 Somnath Banerjee \\
  CSE Department\\
  Jadavpur University, India \\
   \And
 Sudip Kumar Naskar \\
  CSE Department\\
  Jadavpur University, India\\
  \And
   Paolo Rosso \\
  PRHLT Research Center \\
  Universitat Polit\`ecnica de Val\`encia, Spain \\
  \And
  Sivaji Bandyopadhyay \\
  CSE Department\\
  Jadavpur University, India \\
}

\begin{document}
\maketitle
\begin{abstract}
Question classification (QC) is a prime constituent of automated question answering system.
The work presented here demonstrates that the combination of multiple models achieve better classification performance than those obtained with existing individual models for the question classification task in Bengali.
We have exploited state-of-the-art multiple model combination techniques, i.e., ensemble, stacking and voting, to increase QC accuracy. 
Lexical, syntactic and semantic features of Bengali questions are used for four well-known classifiers, namely Na\"{\i}ve Bayes, kernel Na\"{\i}ve Bayes, Rule Induction, and Decision Tree, which serve as our base learners. 
Single-layer question-class taxonomy with 8 coarse-grained classes is extended to two-layer taxonomy by adding 69 fine-grained classes. 
We carried out the experiments both on  single-layer and two-layer taxonomies.  
Experimental results  confirmed that classifier combination approaches outperform  single classifier classification approaches by 4.02\% for coarse-grained question classes. 
Overall, the stacking approach produces the best results for fine-grained classification and achieves 87.79\% of accuracy. 
The approach presented here could be used in other Indo-Aryan or Indic languages to develop a question answering system. 
\end{abstract}


 independently, namely, Na\"{\i}ve Bayes, kernel Na\"{\i}ve Bayes, Rule Induction and Decision Tree. 
Both theoretical \cite{hansen-salamon:1990,krogh-vedelsby:1995} and empirical \cite{hashem:1997,opitz-Shavlik:1996,shavlik:1996} studies confirm that the classifier combination approach is generally more accurate than any of the individual classifiers making up the ensemble. 
Furthermore, a number of studies \cite{Xin:2006,jia:2007} were successfully carried out on classifier combination methods for the QC task which outperformed the individual classifiers.
Therefore, we consider classifier combination  for classifying Bengali questions.
To the best of our knowledge, classifier combination methods have not been employed for the QC task for Indian languages, prior to the work reported in this paper.
As discussed earlier, mainly rule-based approach was employed for the QC task along with individual classifier based approach.
Furthermore, no research work can be found in the literature for fine-grained question classification in Bengali. 
The deep learning framework performs well when  large datasets are available for training and the framework is less effective than traditional machine learning approaches when the training datasets are small in size. 
In this work, we deal with a dataset which has only 1,100 samples. Therefore, we prefer classifier combination approach over deep learning.  
%
Li and Roth \cite{li-roth:2006} and Lee \textit{et al} \cite{lee:2005} proposed 50 and 62 fine grained classes for English and Chinese QC respectively. 
In our work, we proposed 69 fine grained question classes to develop a two-layer taxonomy for Bengali QC.

\section{Proposed Question Taxonomies}
\label{sec:taxonomy}
The set of question categories is referred to as question taxonomy or question ontology. 
Since Bengali question classification is at an early stage of development, for simplicity initially a single-layer taxonomy for Bengali question types was proposed in \cite{sbQA:2012} which consists of only eight coarse-grained classes and no fine-grained classes. 
No other investigation have been carried out for coarse-grained Bengali taxonomies till date. 
Later, based on the coarse-grained classes in \cite{sbQA:2012}, fine-grained question classes were proposed in \cite{sb-paclic:2013}. 
Table~\ref{tab:taxonomy} presents the Bengali Question taxonomy proposed in \cite{sbQA:2012,sb-paclic:2013}.

\begin{table}[h]
	\centering
	\caption{Two-layer Bengali Question Taxonomies}
	\label{tab:taxonomy}       
			\begin{tabular}{p{1.8cm} p{7.3cm}}
				\hline
				Coarse-grained & Fine-grained\\\hline\hline
				Person (PER)	& GROUP, INDIVIDUAL, APPELLATION, INVENTOR/ DISCOVERER, POSITION, OTHER \\
				Organization (ORG)	& BANK, COMPANY, SPORT-TEAM, UNIVERSITY, OTHER \\
				Location (LOC)	& CITY, CONTINENT, COUNTRY, ISLAND, LAKE, MOUNTAIN, OCEAN, ADDRESS, RIVER, OTHER \\
				Temporal (TEM)	& DATE, TIME, YEAR, MONTH, WEEK, DAY, OTHER \\
				Numerical (NUM)	& AGE, AREA, COUNT, LENGTH, FREQUENCY, MONEY, PERCENT, PHONE-NUMBER, SPEED, WEIGHT, TEMPERATURE, OTHER \\
				Method (METH)	& NATURAL, ARTIFICIAL \\
				Reason (REA)	& INSTRUMENTAL, NON-INSTRUMENTAL \\
				Definition (DEF)	& ANIMAL, BODY, CREATION, CURRENCY, FOOD, INSTRUMENT, OTHER, PLANT, PRODUCT, SPORT, SYMBOL, TECHNIQUE, TERM, WORD \\
				Miscellaneous (MISC)	& COLOR, CURRENCY, ENTERTAINMENT, LANGUAGE, OTHER, VEHICLE, AFFAIR, DISEASE, PRESS, RELIGION \\
				\hline
			\end{tabular}
\end{table}

\begin{table} [ht]
	\centering
	\caption{Bengali question examples} 
	\label{tab:taxonomy-examples}
	\begin{tabular}{ p{2cm}  p{12.0cm}}
		Class & Example\\\hline\hline
		Person (PER) &  ke gOdZa prawiRTA karena ?		  
		  
		  (gloss: Who established Goura?)\\ \hline
		
		Organization (ORG) &  sinXu saByawAra safgeV koVna saByawAra mila KuzjeV pAoVyZA yAyZa ?
		
		(gloss: Which civilization has resemblance with the Indus Valley Civilization ?)\\\hline
		Location (LOC) & gOdZa koWAyZa abashiwa ?
		
		(gloss: Where is Goura situated ?)\\ \hline
		
		Temporal (TEMP) & BAikiM-2 kawa baCara karmakRama Cila ?
		
		(gloss:For how many years Vaikin-2 was working ?)\\\hline
		Numerical (NUM) & sUrya WeVkeV Sani graheVra gadZa xUrawba kawa ? 
		
		(gloss: What is the average distance of the planet Saturn from the Sun ?)\\\hline
		Method (METH) &  AryasaByawA mahilArA  kiBAbeV cula bAzXawa ?
		
		(gloss: How do the women braid hair in the Arya Civilization)\\\hline
		Reason (REA) & AryasaByawAkeV keVna bExika saByawA balA hayZa ?
		
		(gloss: Why the Arya Civilization is called the Vedic Civilization?)\\\hline
		Definition (DEF) &  beVxa ki ?
		
		(gloss: What is Veda?)\\\hline
		Miscellaneous (MISC) & Arya samAjeV cArati barNa ki ki Cila ?
		
		(gloss: What are the four classes in the Arya Society?)\\\hline
		\hline 	      
	\end{tabular}
\end{table} 
The taxonomy proposed by Li and Roth \cite{li-roth:2006} contains 6 coarse-grained classes: \textit{Abbreviation, Description, Entity, Human, Location, Numeric}. \textit{Abbreviation} and \textit{Description} classes of \cite{li-roth:2006} are not present in Bengali taxonomy. Two coarse-grained classes of \cite{li-roth:2006}, namely, \textit{Entity} and \textit{Human}  have resemblance with \textit{Miscellaneous} and \textit{Person} respectively in Bengali taxonomy. While \textit{Location} and \textit{Number} classes are present in both the taxonomies, \textit{Organization} and \textit{Method} classes are not present in \cite{li-roth:2006}. In 2-layer Bengali taxonomy, 15 fine-grained classes of \cite{li-roth:2006} are not present, namely, \textit{abbreviation, expression, definition, description, manner, reason, event, letter, substance, title, description, state, code, distance, order}.

All the coarse-grained classes of Lee \textit{et al} \cite{lee:2005} are present in Bengali taxonomy. However, the \textit{Method} class of Bengali taxonomy is not present in \cite{lee:2005}. The \textit{Artifact} class of \cite{lee:2005} is similar to \textit{Definition} and \textit{Miscellaneous} of Bengali-taxonomy. In 2-layer Bengali taxonomy, 9 fine-grained classes of \cite{lee:2005} are not included, namely, \textit{firstperson, planet, province, political system, substance, range, number, range, order}.

The 5 fine-grained classes are introduced in Bengali taxonomy which are not present in \cite{li-roth:2006} and \cite{lee:2005}. The 5 classes are: \textit{AGE, NATURAL, ARTIFICIAL, INSTRUMENTAL, NON- INSTRUMENTAL}. 
The \textit{NATURAL} and \textit{ARTIFICIAL} fine - grained classes belong to \textit{Method} coarse-grained class which is not present in \cite{li-roth:2006} and \cite{lee:2005}. Similarly,  \textit{INSTRUMENTAL, NON-INSTRUMENTAL} fine-grained classes belong to the \textit{Reason} coarse-grained class. Also, the \textit{Reason} coarse-grained class is not present in \cite{li-roth:2006} and \cite{lee:2005}. 
The \textit{AGE} fine-grained class belong to \textit{Numerical} coarse class. 

The taxonomies proposed in \cite{li-roth:2006} and \cite{lee:2005} did not deal with causal and procedural questions. The proposed Bengali 2-layer taxonomy is based on the only available Bengali QA dataset \cite{sbQA:2012} which contains causal and procedural questions. Therefore, the Bengali taxonomy contains question classes of causal and procedural questions. A few fine-grained classes of \cite{li-roth:2006} and \cite{lee:2005} are not included in the taxonomy because such questions are not present in the Bengali QA dataset. However, the proposed Bengali taxonomy is not final for Bengali QA task. Increasing the size of the said dataset is still in the process. Therefore, it is expected that the missing fine-grained classes will be incorporated in the taxonomy in future.  

\section{Features for Question Classification}
\label{sec:features}

In the task of machine learning based QC, deciding the optimal set of features to train the classifiers is crucial. 
The features used for the QC task can be broadly categorized into three different types: lexical, syntactic and semantic features \cite{loni-question:2011}. 
In the present work, we also employed these three types of features suitable for the Bengali QC task.

Loni \textit{et al} \cite{loni-question:2011} represented questions for the QC task similar to document representation in the vector space model, i.e., a question is represented as a vector described by the words inside it. 
Therefore, a question $Q_i$ can be represented as below:

$Q_i = (W_{i1}, W_{i2}, W_{i3}, \dots, W_{i(N-1)}, W_{iN})$
\\where, $W_{ik} =$ frequency of the term $k$ in question $Q_i$, and $N =$ total number of terms. 

Due to the sparseness of the feature vector, only non-zero valued features are kept. Therefore, the size of the samples is quite small despite the huge size of feature space. 
All lexical, syntactic and semantic features can be added to the feature space which expands the feature vector.

In the present study, the features employed for classifying questions (cf. Table \ref{tab:taxonomy}) are described in the following subsections. 
In addition to the features used for the coarse-grained classification, fine-grained classification uses an additional feature, namely coarse-class, i.e. label of the coarse-grained class.

\subsection{Lexical Features}
Lexical features ($f_L$) of a question are extracted from the words appearing in the question. Lexical features include interrogative-words, interrogative-word-positions, interrogative-type, question-length, end-marker and word-shape. 

\paragraph{$\bullet$ \textit{Interrogative-words and interrogative-word positions:}} The interrogative-words (e.g., what, who, which etc.) of a question are important lexical features. They are often referred to as \textit{wh}-words. Huang \textit{et al} \cite{huang:2008,huang:2009} showed that considering question interrogative-word(s) as a feature can improve the performance of question classification task for English QA. Because of the relatively free word-ordering in Bengali, interrogative-words might not always appear at the beginning of the sentence, as in English. Therefore,
the position of the interrogative (wh) words along with the interrogative words themselves have been considered as the lexical features. The position value is based on the appearance of the interrogative word in the question text and it can have any of the three values namely, first, middle and last. 

\paragraph{$\bullet$ \textit{Interrogative-type:}} Unlike in English, there are many interrogatives present in the Bengali language. Twenty six Bengali interrogatives were reported in \cite{sbQA:2012}. In the present work, 
the Bengali interrogative-type (wh-type) is considered as another lexical feature. 
In \cite{sbQA:2012}, the authors concluded that Bengali interrogatives not only provide important information about the expected answers but also indicate the number information (i.e., singular vs plural). 
In \cite{sbQA:2012}, wh-type was classified to three categories:  Simple Interrogative (SI) or Unit Interrogative (UI), Dual Interrogative (UI) and Compound/Composite Interrogative (CI).

\paragraph{$\bullet$ \textit{Question length:}} Blunsom \textit{et al} \cite{blunsom:2006} introduced the length of a question as an important lexical feature which is simply the number of words in a question. We also considered this feature for the present study.

\paragraph{$\bullet$ \textit{End marker:}} The end marker plays an important role in Bengali QC task. Bengali question is end with either `?' or `\textbar'. 
It has been observed from the experimental corpus that if the end marker is `\textbar' (similar to dot (.) in English), then the given question is a definition question.

\paragraph{$\bullet$ \textit{Word shape:}}
The word shape of each question word is considered as a feature.
Word shapes refer to apparent properties of single words. 
Huang \textit{et al} \cite{huang:2008} introduced five categories for word shapes: all digits, lower case, upper case, mixed and other. Word shape alone is not a good feature for QC, however, when it is combined with other kinds of features, it usually improves the accuracy of QC \cite{huang:2008,loni-question:2011}. Capitalization feature is not present in Bengali; so we have considered only the other three categories, i.e., all digits, mixed and other.

Example-1: \textit{ke gOdZa prawiRTA karena ?}\footnotetext{All the Bengali examples in this paper are written in WX \cite{wx-notation:2010} notation which is a transliteration scheme for representing Indian languages in ASCII.}

Gloss: Who established Goura?

Lexical features: wh-word: ke; wh-word position: first; wh-type: SI; 
question length: 5; end marker: ? word shape: other 

\subsection{Syntactic Features}

Although different works extracted several syntactic features ($f_S$), the most commonly used $f_S$ are Part of Speech (POS) tags and head words \cite{loni-survey:2011}.

\paragraph{$\bullet$ \textit{POS tags:}} In the present work, we used the POS tag of each word in a question such as NN (Noun), JJ (adjective), etc. POS of each question word is added to the feature vector. 
A similar approach was successfully used for English \cite{li-roth:2006,blunsom:2006}. 
This feature space is sometimes referred to as the bag-of-POS tags  \cite{loni-question:2011}.
The Tagged-Unigram (TU) feature was formally introduced by \cite{loni-question:2011}. 
TU feature is simply the unigrams augmented with POS tags. Loni \textit{et al} \cite{loni-question:2011} showed that considering the tagged-unigrams instead of normal unigrams can help the classifier to distinguish a word with different tags as two different features.
For extracting the POS tags, the proposed classification work in Bengali uses a \textit{Bengali Shallow Parser}\footnote{ http:// ltrc.iiit.ac.in/analyzer/bengali/} which produces POS tagged data as intermediate result.

\paragraph{$\bullet$ \textit{Question head word:}} Question head-word is the most informative word in a question as it specifies the object the question is looking for \cite{huang:2008}. Correctly identifying head-words can significantly improve the classification accuracy. 
For example, in the question ``\textit{What is the oldest city in Canada?}'' the headword is `city'. The word `city' in this question can highly contribute to classify this question as LOC: city.

Identifying the question's head-word is very challenging in Bengali because of its syntactic nature and no research has been conducted so far on this. Based on the position of the interrogative in the question, we use heuristics to identify the question head-words. According to  the position of the interrogative, three cases are possible. 
\\-- \textit{Position-I (at the beginning):} If the question-word (i.e., marked by WQ tag) appears at the beginning then the first NP chunk after the interrogative-word is considered as the head-word of the question. 
Let us consider the following question.

Example-2: \textit{ke}(/WQ) \textit{gOdZa}(/NNP) \textit{prawiRTA}(/NN)\\ \textit{karena}(/VM) ?(/SYM)

English Gloss: Who established Goura ?

In the above example, gOdZa is the head-word. 
\\-- \textit{Position-II (in between)}: If the position of the question-word is neither at the beginning or at the end then the immediate NP-chunk before the interrogative-word is considered as the head-word. 
Let us consider the following question.

Example-3: \textit{gOdZa}(/NNP) \textit{koWAyZa}(/WQ) \textit{abashiwa}(/JJ) ?(/SYM)

English Gloss: Where is Goura situated ?

In the above example gOdZa is considered as the question head-word.
\\-- \textit{Position-III (at the end)}: If the question-word appears at the end (i.e., just before the end of sentence marker) then the immediate NP-chunk before the interrogative-word is considered as the question head-word. Therefore, a similar action is taken for Position II and III. 

Example-4:[\textit{bAMlAxeSe arWanIwi kaleja}](/NNP) \textit{kayZati} (/WQ) ?(/SYM)

English Gloss: How many economics colleges are in \\Bangladesh?

Therefore, in the Example-4 [\textit{bAMlAxeSe arWanIwi kaleja} ] is the question head-word. 

Now, if we consider the example ``\textit{ke gOdZa prawiRTA karena ?}'' then the syntactic features will be: [\{WQ, 1\},\{NNP, 1\}, \{NN, 1\}, \{VM, 1\},\{head-word,gOdZa\}]. 
Here a feature is represented as \{$\langle$ POS, frequency $\rangle$\}.

\subsection{Semantic Features}  

Semantic features ($f_M$) are extracted based on the semantics of the words in a question. 
In this study, related words and named entities are used as $f_M$.

\paragraph{$\bullet$ \textit{Related word:}} 
A Bengali synonym dictionary is used to retrieve the related words. Three lists of related words were manually prepared by analyzing the training data. 

date:\{ \textit{janmaxina, xina, xaSaka, GantA, sapwAha, mAsa, baCara,} ...,etc.\}; 

food:\{ \textit{KAbAra, mACa, KAxya, mAKana, Pala,Alu, miRti, sbAxa,} ..., etc.\}; 

human authority:\{ \textit{narapawi, rAjA, praXAnamanwrI, \\bicArapawi, mahAparicAlaka, ceyZAramyAna, jenArela, sulawAna, samrAta, mahAXyakRa,} ..., etc.\}; 

If a question word belongs to any of the three lists (namely date, food, human activity), then its category name is added to the feature vector. 
For instance, the question ``\textit{ke gedZera sbAXIna narapawi Cilena ?}'' (gloss: who was the independent ruler of Goura ?) contains the word \textit{narapawi} which belongs to the human authority list.
For this example question the semantic feature is added to the feature vector as: [\{human-authority, 1\}].

\paragraph{$\bullet$ \textit{Named entities:}} We used named entities (NE) as a semantic feature which was also recommended in other works \cite{li-roth:2006,blunsom:2006} on other languages. 
To identify the Bengali named entities in the question text, a Margin Infused Relaxed Algorithm (MIRA) based Named Entity Recognizer (NER) \cite{sb-mira:2014} is used for the present study.
For the Example-5 question, the NE semantic feature is added to the feature vector as: [{Location, 1}]. 

Example-5: \textit{ke gOdZa}[Location] \textit{prawiRTA karena}? 

English Gloss: Who established Goura ?

\section{Experiments and Results}
\label{sec:experiment-results}

Many supervised learning approaches \cite{huang:2008,blunsom:2006,li:2002} have been proposed for QC over the years. 
But these approaches primarily differ in the classifier they use and the features they train their classifier(s) on \cite{loni-survey:2011}. 
We assume that a Bengali question is unambiguous, i.e., a question belongs to only one class. 
Therefore, we considered multinomial classification which assigns the most likely class from the set of classes to a question. 
Recent studies \cite{zhang-lee:2003,huang:2008,silva:2011} also considered one label per question.

We used state-of-the-art classifier combination approaches: ensemble, stacking and voting. 
We have used two contemporary methods for creating accurate ensembles, namely, bagging and boosting. 
We employed the Rapid Miner tool for all the experiments reported here. 
Each of the three classifier combination approaches was tested with Na\"{\i}ve Bayes (NB), Kernel Na\"{\i}ve Bayes (k-NB), Rule Induction (RI) and Decision Tree (DT) classifiers.

Classification accuracy is used to evaluate the results of our experiments.
Accuracy is the widely used evaluation metric to determine the class discrimination ability of classifiers, and is calculated using the following equation:
\[ 
\mbox{accuracy} = \frac{\mbox{number of correctly classified samples}}{\mbox{total number of tested samples}}
\]

\subsection{Corpus Annotation and Statistics}
\label{sec:dataset}

We carried out our experiments on the dataset described in \cite{sbQA:2012}. 
The questions in this dataset are acquired from different domains, e.g., education, geography, history, science, etc.
We hired two native language (i.e., Bengali)  specialists for annotating the corpus. Another senior native language expert was hired to support the two language specialists. The annotators were instructed to consult the  senior native language expert in case of any confusion.
In order to minimize disagreement, two language specialists gathered to discuss the question taxonomy in detail before initiating the annotation task. 
We set a constraint that a question will be annotated such that it is unambiguous, i.e., only a question class will be assigned to a question.  
We measured the inter-annotator agreement using non-weighted kappa coefficients \cite{kappa:1960}. The kappa coefficient for the annotation task was 0.85 which represents very high agreement.
In case of or disagreement, the senior language specialist took the final decision.
\begin{table}[htbp]
	\centering
	\caption{Corpus statistics}
    \label{tab:dataset_statistics}%
	\begin{tabular}{lcccc}
		\hline
		\textbf{Class} & \textbf{Train} & \textbf{Test} & \textbf{Overall} \\
		\hline
    Person   & 172   & 90    & 262 \\
    Organization   & 74    & 30    & 104 \\
    Location   & 76    & 30    & 106 \\
    Temporal   & 81    & 35    & 116 \\
    Numerical   & 71    & 30    & 101 \\
    Methodical   & 75    & 29    & 104 \\
    Reason   & 73    & 26    & 99 \\
    Definition   & 78    & 38    & 116 \\
    Miscellaneous   & 69    & 23    & 92 \\\hline
    Total  & 769   & 331   & 1100 \\
	\end{tabular}%
\end{table}%

The class-specific distribution of questions in the corpus is given in Table~\ref{tab:dataset_statistics}.
It can be observed from Table~\ref{tab:dataset_statistics} that the most frequent question class in the dataset is `Person'. 
The dataset contains a total of 1,100 questions.
We divided the question corpus into 7:3 ratio for experimentation. 
The experimental dataset consists of 1100 Bengali questions of which 70\% are used for training and the rest (331 questions, 30\%) for testing the classification models.

\subsection{Coarse-Grained Classification}
The empirical study of state-of-the-art classifier combination approaches (i.e., ensemble, stacking, and voting) was performed on the said dataset using four classifiers - namely, NB, k-NB, RI and DT. 
Each experiment can be thought of as a combination of three experiments since each classifier model was tested on \{$f_L$\}, $\{f_L,f_S\}$ and $\{f_L,f_S,f_M\}$ feature sets separately. 
Overall thirteen experiments were performed for coarse-grained classification and the evaluation results are reported in Table~\ref{tab:coarse-results1}.
\begin{table*}[t]
	\centering
	\caption{Classifier combination results for coarse-grained classification}
	\label{tab:coarse-results1}       
	\begin{tabular}{lccccc}
		\textbf{Approach}         & \textbf{Base-Learner} & \multicolumn{1}{c}{\textbf{Model-Learner}} & \textbf{$f_L$} & $f_L$+ $f_S$ & $f_L$+$f_S$+$f_M$ \\ \hline
		\multirow{4}{*}{Bagging}  & NB                    & x                                          & 81.53        & 82.77             & 83.25                \\
		& k-NB                  & x                                          & 82.09        & 83.37             & 84.22                \\
		& RI                    & x                                          & 83.96        & 85.61             & 86.90                \\
		& DT                    & x                                          & 85.23        & 86.41             & \textbf{91.27}       \\ \hline
		\multirow{4}{*}{Boosting} & NB                    & x                                          & 81.74        & 82.71             & 83.51                \\
		& k-NB                  & x                                          & 83.86        & 85.63             & 86.87                \\
		& RI                    & x                                          & 83.55        & 85.59             & 86.27                \\
		& DT                    & x                                          & 85.21        & 86.58             & \textbf{91.13}       \\ \hline
		\multirow{4}{*}{Stacking} & k-NB, RI, DT          & NB                                         & 81.76        & 82.79             & 83.64                \\
		& NB, RI, DT            & k-NB                                       & 83.86        & 85.54             & 86.75                \\
		& NB, k-NB, DT          & RI                                         & 85.55        & 87.69             & \textbf{91.32}       \\
		& NB, k-NB, RI,         & DT                                         & 85.07        & 86.73             & 89.13                \\\hline
		Voting                    & NB, k-NB, RI, DT      & x                                          & 86.59        & 88.43             & \textbf{91.65} \\ \hline      
	\end{tabular}
\end{table*}

\subsubsection{Ensemble Bagging}
\label{sec:coarse-bagging}
\begin{figure} [b]
	\centering
	\includegraphics[scale= 0.5]{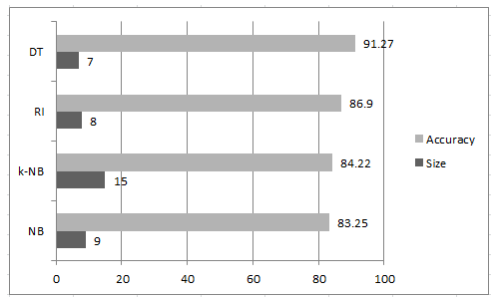}
	\caption{Size and Accuracy variation in Bagging with $\{f_L,f_S,f_M\}$}
	\label{fig:bagging-coarse}       
\end{figure}
The bagging approach was applied separately to four classifiers (i.e., NB, k-NB, RI and DT) and the obtained accuracies are summarized in Table~\ref{tab:coarse-results1}. 
Initially, the size (i.e., number of iterations) of the base learner was set to 2. 
Subsequently, experiments were performed with gradually increasing size ($size>2$). The classification accuracy enhanced with increase in size. 
However, after a certain size, the accuracy was almost stable. At $size=2$ and feature set $\{f_L,f_S,f_M\}$, the NB classifier achieved 82.23\% accuracy and at $size\geq9$, it became stable with 83.25\% accuracy. 
At $size=2$ and feature set $\{f_L,f_S,f_M\}$,the k-NB classifier achieved 83.87\% accuracy and at $size\geq15$, it became stable with 84.22\% accuracy. 
At $size=2$ and feature set $\{f_L,f_S,f_M\}$, the RI classifier achieved 85.97\% accuracy and at $size\geq8$, it became stable with 86.90\% accuracy. 
At $size=2$ and feature set $\{f_L,f_S,f_M\}$,the DT classifier achieved 88.09\% accuracy and at $size\geq7$, it became stable with 91.27\% accuracy.  
It was observed from the experiments that with bagging the DT classifier performs best on any feature set for any size.
For the experiments with the $f_L$ features, the bagging size of NB, k-NB, RI and DT are 12, 19, 11 and 10 respectively after which classification accuracy becomes stable. 
Similarly, for experiments with $\{f_L,f_S\}$ feature set, the optimal bagging sizes are 10, 17, 9 and 8 for NB, k-NB, RI and DT respectively after which the corresponding classification accuracies converge.
The Figure~\ref{fig:bagging-coarse} shows the variation in size and accuracy for the best feature set.

\subsubsection{Ensemble Boosting}
\label{sec:coarse-boosting}

\begin{figure} [h]
	\centering
	\includegraphics[scale= 0.5]{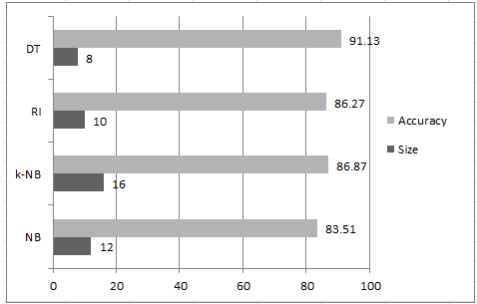}
	\caption{Size and Accuracy variation in Boosting with $\{f_L,f_S,f_M\}$}
	\label{fig:boosting-coarse}       
\end{figure}

Like bagging, boosting (AdaBoost.M1) was also applied separately to the four base classifiers. Table~\ref{tab:coarse-results1} tabulates the accuracies obtained with the boosting approach with the four classifiers. Here, we empirically fixed the iterations of boosting for the four classifiers to 12, 16, 10 and 8 respectively for the feature set $\{f_L,f_S,f_M\}$, since the corresponding weight of $\frac{1}{{\beta}_t}$ becomes less than 1 beyond those values. If $\frac{1}{{\beta}_t}$ is less than 1, then the weight of the classifier model in boosting may be less than zero for that iteration. The Figure~\ref{fig:boosting-coarse} shows the variation in size and accuracy for the best feature set.

Similarly, for the feature sets $\{f_L,f_S\}$ and \{$f_L$\} 
the iterations are set to 13, 18, 12, 9 and 14, 19, 14, 11 respectively for the four classifiers.  Overall the DT classifier performs the best. However, unlike in bagging, k-NB performs better than RI with boosting.

\subsubsection{Stacking}
\label{sec:coarse-stacking}

In stacking, three out of the four classifiers are used as the base learners (BL) and the remaining classifier is used as the model learner (ML). Therefore, four experiments were conducted separately for each of the four classifiers as the ML. The obtained accuracies are summarized in Table~\ref{tab:coarse-results1}.

Experimental results revealed that with RI as the model learner and NB, k-NB, DT as the base learners, the classifier achieves the best classification accuracy.

\subsubsection{Voting}
\label{sec:coarse-voting}

In voting, four classifiers altogether were used as the base learners and majority vote was used as voting approach. The evaluation results of the voting approach are presented in Table~\ref{tab:coarse-results1}.

\subsection{Result Analysis of Coarse-Grained Classification}

Classifier combination is an established research known under different names in the literature: committees of learners, mixtures of experts, classifier ensembles, multiple classifier systems, etc. A number of research \cite{breiman-bagging:1996,clemen:1989,hansen-salamon:1990,hashem:1997} established that classifier combination could produce better results than single classifier. Generally, the key to the success of classifier combination approach is that it builds a set of diverse classifiers where each classifier is based on different subsets of the training data. Therefore, our objective is to verify the impact of the classifier combination approaches over the individual classifier approaches on Bengali QC task.

The automated Bengali QC system by \cite{sbQA:2012} is based on four classifiers, namely NB, k-NB, RI and DT, which were used separately. 
\begin{table}[h]
	\centering
	\caption{Experimental results of \cite{sbQA:2012}}
	\label{tab:coarse-results2}       
	\begin{tabular}{lccc}
		
		Classifier & $f_L$ & $f_L$+$f_S$ &  $f_L$+$f_S$+$f_M$ \\ 
		\hline 
		NB	& 80.65	& 81.34	& 81.89 \\
		k-NB &	81.09	& 82.37	& 83.21 \\
		RI	& 83.31	& 84.23	& 85.57 \\
		DT	& 84.19	& 85.69	& 87.63 \\ \hline 
	\end{tabular} 
\end{table}

The experimental results obtained by \cite{sbQA:2012} are shown in Table~\ref{tab:coarse-results2}. 
In that work, NB was used as the baseline and the DT classifier achieved the highest accuracy of 87.63\% (cf. Table~\ref{tab:coarse-results2}). 
A comparison of the results in Table~\ref{tab:coarse-results1} and Table~\ref{tab:coarse-results2} reveals that each classifier combination model performs better than the single classifier models in terms of classification accuracy. 
The prime reason is that classifier combination approaches reduce model bias and variance more effectively than individual classifiers.

In comparison to the earlier experiments reported in \cite{sbQA:2012}, with the bagging approach,  classification accuracy of each classifier increases notably with bagging.
The classification accuracy on the \{$f_L$\}, $\{f_L,f_S\}$ and $\{f_L,f_S,f_M\}$ feature sets increases by 1.04\%, 0.72\% and 3.64\% for best performing DT classifier.
Similarly, with the boosting approach, {the classification accuracy for the best performing DT classifiers notably increases by 1.02\%, 0.89\% and 3.50\% on \{$f_L$\}, $\{f_L,f_S\}$ and $\{f_L,f_S,f_M\}$ feature set}. 
The stacking approach increases the accuracy on the $\{f_L,f_S\}$ feature set than the bagging and boosting approaches. 
This approach increases the classification accuracy by 1.36\%, 2.74\% and 0.69\% on the \{$f_L$\}, $\{f_L,f_S\}$ and $\{f_L,f_S,f_M\}$ feature sets respectively. 
The voting approach not only increases the classification accuracy, but also provides the maximum accuracy for all the feature sets than the other combined approaches. The voting approach increases the classification accuracy on the \{$f_L$\}, $\{f_L,f_S\}$ and $\{f_L,f_S,f_M\}$ feature sets by 2.40\%, 2.40\% and 4.02\% respectively.
Therefore, overall the voting approach with majority voting performed the best among the four classifier combination approaches.

Bagging approach helps to avoid over fitting by reducing variance \cite{breiman-bagging:1996}. However, after certain iteration, it cannot reduce variance. Hence, after certain iteration, it does not improve the performance of the model. Therefore, we observed that after size (i.e., number of iterations), it was unable to enhance the accuracy.

On the other hand, boosting approach enhance the performance of the model by primarily reducing the bias \cite{schapire-boosting:1990}. However, after certain iteration (size) it cannot be improved. Because after certain iterations, the corresponding weight of $\frac{1}{{\beta}_t}$ becomes less than 1. If $\frac{1}{{\beta}_t}$ is less than 1, then the weight of the classifier model in boosting may be less than zero for that iteration. Therefore, we were not able to improve the accuracy after specific boosting size.

In stacking, the model learner is trained on the outputs of the base learners that are trained based on a complete training set \cite{wolpert-stacking:1992}. Out experiment reveals that RI as model learner and NB, k-NB, DT as the base learners outperforms the other models.

In the context of Bengali question classification task, we conclude from the experimental results that although classifier combination approach outperforms the individual classifier approach, the impact of different classifier combination approaches is almost same for the Bengali course classes. Because, we obtained almost similar accuracy for different classifier combination approaches, namely, ensemble, stacking and voting.   

\begin{table}[h]
	\centering
	\caption{Fine-grained classification using individual classifiers}
	\label{tab:fine-results-individual} 
		\begin{tabular}{llccc}
			Classifier & Class & $f_L$ & $f_L$+$f_S$ &  $f_L$+$f_S$+$f_M$ \\ \hline
			\multirow{9}{*}{NB} &  $F_{PER}$   & 74.07 & 75.54 & 77.07 \\
			&  $F_{ORG}$   & 75.33 & 76.55 & 77.70 \\
			&  $F_{LOC}$   & 76.15 & 77.02 & 77.87 \\
			&  $F_{TEM}$   & 75.74 & 77.16 & 77.97 \\
			&  $F_{NUM}$   & 74.61 & 75.45 & 76.55 \\
			&  $F_{METH}$   & 76.35 & 77.42 & 78.50 \\
			&  $F_{REA}$   & 76.19 & 77.20 & 78.02 \\
			&  $F_{DEF}$   & 76.30 & 77.45 & 78.56 \\
			&  $F_{MISC}$   & 75.80 & 76.95 & 77.40 \\ \hline
			
			\multirow{9}{*}{k-NB} &  $F_{PER}$   & 75.72 & 77.33 & 78.41 \\
			&  $F_{ORG}$   & 76.76 & 77.97 & 79.28 \\
			&  $F_{LOC}$   & 77.52 & 78.55 & 79.40 \\
			&  $F_{TEM}$   & 77.22 & 78.73 & 79.57 \\
			&  $F_{NUM}$   & 76.09 & 76.94 & 78.05 \\
			&  $F_{METH}$   & 77.92 & 79.14 & 80.24 \\
			&  $F_{REA}$   & 77.82 & 79.36 & 80.33 \\
			&  $F_{DEF}$   & 77.99 & 79.40 & 80.43 \\
			&  $F_{MISC}$   & 77.37 & 78.74 & 79.60 \\ \hline
			
			\multirow{9}{*}{RI} &  $F_{PER}$   & 77.96 & 79.04 & 80.12 \\
			&  $F_{ORG}$   & 78.29 & 79.56 & 80.75 \\
			&  $F_{LOC}$   & 77.67 & 78.36 & 79.18 \\
			&  $F_{TEM}$   & 79.17 & 80.76 & 81.73 \\
			&  $F_{NUM}$   & 78.04 & 79.03 & 80.42 \\
			&  $F_{METH}$   & 79.87 & 81.00 & 82.12 \\
			&  $F_{REA}$   & 79.62 & 80.93 & 82.06 \\
			&  $F_{DEF}$   & 78.98 & 80.28 & 81.28 \\
			&  $F_{MISC}$   & 78.59 & 79.91 & 80.90 \\ \hline
			
			\multirow{9}{*}{DT} &  $F_{PER}$   & 80.37 & 82.06 & 83.61 \\
			&  $F_{ORG}$   & 78.78 & 80.26 & 81.68 \\
			&  $F_{LOC}$   & 78.51 & 79.63 & 80.94 \\
			&  $F_{TEM}$   & 80.58 & 82.03 & 83.50 \\
			&  $F_{NUM}$   & 79.00 & 80.50 & 81.85 \\
			&  $F_{METH}$   & 80.62 & 82.55 & 84.47 \\
			&  $F_{REA}$   & 80.51 & 82.49 & 84.42 \\
			&  $F_{DEF}$   & 79.89 & 81.07 & 82.49 \\
			&  $F_{MISC}$   & 79.74 & 81.72 & 84.07 \\ \hline
		\end{tabular}
\end{table}

\subsection{Fine-Grained Classification}
Initially, we applied NB, k-NB, RI and DT classifiers separately. 
Each classifier was trained with \{$f_L$\}, 
$\{f_L,f_S\}$ and $\{f_L,f_S,f_M\}$ feature sets. 
\begin{table*}[h]
	\centering
	\caption{Ensemble results of fine-grained classification}
	\label{tab:fine-grained-ensemble}       
			\begin{tabular}{llccc||ccc}
				& &   \multicolumn{3}{c||}{\textbf{Bagging}} &  \multicolumn{3}{c}{\textbf{Boosting}} \\ \cline{3-8}
				& & $f_L$ & $f_L$+$f_S$ &  $f_L$+$f_S$+$f_M$ &  $f_L$ & $f_L$+$f_S$ &  $f_L$+$f_S$+$f_M$ \\ \cline{3-8}
				\multirow{9}{*}{NB} & $F_{PER}$ & 79.65 & 81.23 & 82.87 & 79.89 & 81.41 & 82.95 \\
				& $F_{ORG}$ & 81.01 & 82.32 & 83.55 & 81.65 & 82.73 & 83.98 \\ 
				& $F_{LOC}$ & 81.89 & 82.82 & 83.73 & 82.28 & 83.85 & 85.04 \\
				& $F_{TEM}$ & 81.45 & 82.97 & 83.84 & 81.89 & 83.01 & 83.97 \\
				& $F_{NUM}$ & 80.23 & 81.13 & 82.31 & 81.02 & 81.92 & 83.03 \\
				& $F_{METH}$ & 82.10 & 83.25 & 84.41 & 82.25 & 83.37 & 84.53 \\
				& $F_{REA}$ & 81.93 & 83.02 & 84.17 & 82.06 & 83.11 & 84.23 \\
				& $F_{DEF}$ & 82.05 & 83.29 & 84.47 & 82.09 & 83.32 & 84.56 \\
				& $F_{MISC}$ & 81.51 & 82.75 & 83.23 & 81.62 & 82.79 & 83.75 \\ \hline
				
				\multirow{9}{*}{k-NB} & $F_{PER}$ & 80.13 & 81.83 & 82.97 & 80.17 & 81.91 & 83.02 \\
				& $F_{ORG}$ & 81.23 & 82.51 & 83.89 & 81.29 & 82.63 & 83.91 \\
				& $F_{LOC}$ & 82.03 & 83.12 & 84.02 & 82.10 & 83.17 & 84.09 \\
				& $F_{TEM}$ & 81.71 & 83.31 & 84.20 & 81.79 & 83.39 & 84.28 \\
				& $F_{NUM}$ & 80.52 & 81.42 & 82.59 & 80.63 & 81.58 & 82.69 \\
				& $F_{METH}$ & 82.45 & 83.75 & 84.91 & 82.48 & 83.79 & 84.98 \\
				& $F_{REA}$ & 82.35 & 83.98 & 85.01 & 82.41 & 84.02 & 85.09 \\
				& $F_{DEF}$ & 82.53 & 84.02 & 85.11 & 82.61 & 84.12 & 85.13 \\
				& $F_{MISC}$ & 81.87 & 83.32 & 84.23 & 81.91 & 83.39 & 84.28 \\ \hline
				
				\multirow{9}{*}{RI} & $F_{PER}$ & 81.85 & 82.98 & 84.12 & 81.92 & 83.06 & 84.22 \\
				& $F_{ORG}$ & 82.19 & 83.53 & 84.78 & 82.25 & 83.61 & 84.85 \\
				& $F_{LOC}$ & 81.54 & 82.27 & 83.13 & 81.55 & 82.26 & 83.15 \\
				& $F_{TEM}$ & 83.12 & 84.79 & 85.81 & 83.18 & 84.85 & 85.93 \\
				& $F_{NUM}$ & 81.93 & 82.97 & 84.43 & 82.01 & 83.03 & 84.49 \\
				& $F_{METH}$ & 83.85 & 85.04 & 86.22 & 83.91 & 85.06 & 86.31 \\
				& $F_{REA}$ & 83.59 & 84.97 & 86.15 & 83.68 & 85.11 & 86.33 \\
				& $F_{DEF}$ & 82.92 & 84.28 & 85.33 & 82.95 & 84.32 & 85.41 \\
				& $F_{MISC}$ & 82.51 & 83.89 & 84.93 & 82.57 & 83.93 & 84.98 \\ \hline
				
				\multirow{9}{*}{DT} & $F_{PER}$ & 84.79 & 86.57 & 88.21 & 84.81 & 86.63 & 88.53 \\
				& $F_{ORG}$ & 83.11 & 84.67 & 86.17 & 83.14 & 84.73 & 86.23 \\
				& $F_{LOC}$ & 82.83 & 84.01 & 85.39 & 82.87 & 84.13 & 85.52 \\
				& $F_{TEM}$ & 85.01 & 86.54 & 88.09 & 85.03 & 86.58 & 88.15 \\
				& $F_{NUM}$ & 83.34 & 84.92 & 86.35 & 83.38 & 84.97 & 86.44 \\
				& $F_{METH}$ & 85.05 & 87.09 & 89.11 & 85.09 & 87.14 & 89.12 \\
				& $F_{REA}$ & 84.93 & 87.02 & 89.06 & 84.96 & 87.11 & 89.09 \\
				& $F_{DEF}$ & 84.28 & 85.53 & 87.02 & 84.29 & 85.55 & 87.05 \\
				& $F_{MISC}$ & 84.12 & 86.21 & 88.69 & 84.15 & 86.23 & 88.73 \\ \hline
			\end{tabular}
\end{table*}
The performance of the classifiers increases gradually with incorporation of syntactic and semantic features (i.e., $\{f_L\} \rightarrow \{f_L,f_S\} \rightarrow  \{f_L,f_S,f_M\}$). 
The NB classifiers achieved around 77\% of accuracy while the k-NB and RI classifiers achieved around 80\% of accuracy for the fine-grained question classes. 
Only the DT classifier obtained more than 80\% accuracy for all the question classes. 
The detailed evaluation results of the fine-grained question classification task using individual classifier are given in Table~\ref{tab:fine-results-individual}. 
The subsequent sections describe the experiments with classifier combination approaches.    

\subsubsection{Ensemble Bagging}
\label{sec:fine-bagging}

In this approach, we use four classifiers as base learners individually: NB, k-NB, RI and DT.
Initially, the base learners are trained using the lexical features ($f_L$). 
Then semantic and syntactic features are added gradually for classification model generation. 
Therefore, three classification models were generated for each base learner. 
Thus, altogether 12 models were prepared for bagging.
\begin{figure} [h]
	\centering {
		\includegraphics[scale= 0.5]{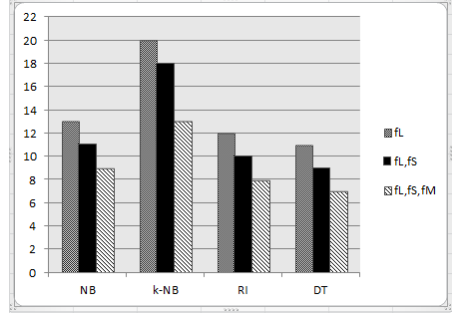}
	}
	\caption{Size variation in Bagging}
	\label{fig:1}       
\end{figure}
Like coarse-grained classification, initially the size (number of iteration) of the base learner was set to 2. 
Subsequently experiments were performed with gradually increasing sizes ($size > 2$).
The classification accuracy increased with higher values of size. 
However, after certain iterations the accuracy was almost stable. 
For the fine-grained classes of PER coarse-class (i.e., $F_{PER}$), with $\{f_L,f_S,f_M\}$) feature set at $size=2$ , the NB classifier achieved 81.98\% classification accuracy and at $size\geq9$, it became stable with 82.87\% accuracy. 
	Similarly, with $\{f_L,f_S,f_M\}$ feature set the k-NB, RI and DT classifiers achieved stable accuracies at $size$ equal to 13, 8 and 7 respectively.
For the lexical feature set, the bagging size of NB, k-NB, RI and DT were 13, 20, 12 and 11 respectively after which the classification accuracy became stable. 
For the combined lexical and syntactic features, the recorded bagging size of NB, k-NB, RI and DT were 11, 18, 10 and 9 respectively. 
Figure~\ref{fig:1} depicts the iteration size for the bagging approach.

\subsubsection{Ensemble Boosting}
\label{sec:fine-boosting}

Like the ensemble bagging approach, we applied boosting (i.e., AdaBoost.M1) separately to the four classifiers. 
Experimental results confirm that performances of the four base classifiers improve slightly using AdaBoost.M1. 
Table~\ref{tab:fine-grained-ensemble} \\presents the results of the boosting experiments and shows that altogether DT outperforms the other classifiers in the ensemble approach, i.e., bagging and boosting.

\begin{figure} [h]
	\centering{
	\includegraphics[scale= 0.5]{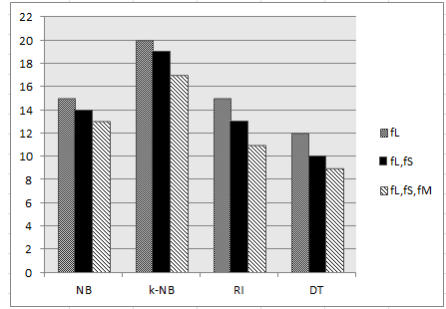}
	}
	\caption{Size variation in Boosting}
	\label{fig:2}       
\end{figure}

In the boosting approach, the number of iterations depends on $\frac{1}{{\beta}_t}$. 
When the value of $\frac{1}{{\beta}_t}$ becomes less than 1, then for that iteration the weight of the boosting classification may be less than zero. 
Hence, we empirically fixed the iterations of AdaBoost.M1 for the four classifiers (i.e., NB, k-NB, RI and DT) to 13, 17, 11 and 9 respectively for the feature set $\{f_L,f_S,f_M\}$ since the weight of $\frac{1}{{\beta}_t}$ becomes less than 1 after those values. 
Similarly, for feature set $\{f_L,f_S\}$ and \{$f_L$\}the iterations were 14, 19, 13, 10 and 15, 20, 15, 12 respectively for the four base classifiers. 
Figure~\ref{fig:2} depicts the iteration sizes of the four classifiers in the boosting approach.

\subsubsection{Stacking}
\label{sec:fine-stacking}

As discussed in Section~\ref{sec:coarse-stacking}, in stacking one classifier plays the role of ML while the remaining classifiers act as BLs.
Therefore, with four classifiers four experiments were conducted separately. 
The obtained accuracies are reported in Table~\ref{tab:fine-grained-stacking}. 
From the experimental results it was observed that the model trained with DT as the model learner and NB, k-NB, RI as the base learners
achieved the best classification accuracy.

\begin{table} [h]
	\centering
	\caption{Results of fine-grained classification with stacking}
	\label{tab:fine-grained-stacking}       
			\begin{tabular}{lllccc}
				\textbf{Base Learner} & \textbf{Model Learner} & \textbf{Class} & $f_L$ & $f_L$+$f_S$ &  $f_L$+$f_S$+$f_M$ \\ \hline
				\multirow{9}{*}{k-NB,RI,DT} & \multirow{9}{*}{NB} &  $F_{PER}$  & 79.81  & 81.67  & 82.86  \\
				& & $F_{ORG}$  & 81.79  & 83.02  & 84.02  \\
				& & $F_{LOC}$  & 81.97  & 83.74  & 84.91  \\
				& & $F_{TEM}$  & 81.45  & 82.81  & 83.73  \\
				& & $F_{NUM}$  & 81.83  & 82.07  & 83.54  \\
				& & $F_{METH}$  & 82.15  & 83.13  & 84.09  \\
				& & $F_{REA}$  & 82.24  & 83.36  & 84.42  \\
				& & $F_{DEF}$  & 81.76  & 83.05  & 84.23  \\
				& & $F_{MISC}$ & 80.21  & 82.33  & 83.21  \\ \hline
				
				\multirow{9}{*}{NB,RI,DT} & \multirow{9}{*}{k-NB} &  $F_{PER}$  & 79.93  & 81.79  & 83.03  \\
				& & $F_{ORG}$  & 81.86  & 83.16  & 84.13  \\
				& & $F_{LOC}$  & 82.08  & 83.82  & 85.06  \\
				& & $F_{TEM}$  & 81.52  & 83.01  & 83.87  \\
				& & $F_{NUM}$  & 81.97  & 82.18  & 83.71  \\
				& & $F_{METH}$  & 82.28  & 83.20  & 84.18  \\
				& & $F_{REA}$  & 82.31  & 83.43  & 84.45  \\
				& & $F_{DEF}$  & 81.82  & 83.21  & 84.31  \\
				& & $F_{MISC}$  & 80.29  & 82.42  & 83.35  \\ \hline
				
				\multirow{9}{*}{NB,k-NB,DT} & \multirow{9}{*}{RI} &  $F_{PER}$  & 80.56  & 83.06  & 84.22  \\
				& & $F_{ORG}$  & 82.86  & 83.98  & 85.03  \\
				& & $F_{LOC}$  & 80.23  & 81.49  & 82.95  \\
				& & $F_{TEM}$  & 83.21  & 84.78  & 85.97  \\
				& & $F_{NUM}$  & 82.37  & 83.42  & 84.77  \\
				& & $F_{METH}$  & 83.54  & 84.93  & 86.27  \\
				& & $F_{REA}$  & 84.03  & 85.75  & 86.73  \\
				& & $F_{DEF}$  & 80.01  & 82.33  & 84.21  \\
				& & $F_{MISC}$  & 82.45  & 83.86  & 84.87  \\ \hline
				
				\multirow{9}{*}{NB,k-NB,RI} & \multirow{9}{*}{DT} &  $F_{PER}$  & 84.97  & 86.69  & 88.71  \\
				& & $F_{ORG}$  & 83.32  & 85.06  & 87.43  \\
				& & $F_{LOC}$  & 82.93  & 84.21  & 85.71  \\
				& & $F_{TEM}$  & 84.84  & 86.13  & 87.95  \\
				& & $F_{NUM}$  & 83.57  & 85.17  & 87.49  \\
				& & $F_{METH}$  & 84.85  & 86.91  & 88.56  \\
				& & $F_{REA}$  & 84.69  & 86.78  & 88.29  \\
				& & $F_{DEF}$  & 84.38  & 85.65  & 87.51  \\
				& & $F_{MISC}$  & 84.02  & 86.11  & 88.42  \\ \hline
			\end{tabular}
\end{table}

\subsubsection{Voting}
\label{sec:fine-voting}
Unlike the ensemble approach, in the voting approach all the classifiers were applied at the same time to predict the question class. 
Table~\ref{tab:fine-grained-voting} tabulates the the accuracies obtained with this approach.

\begin{table}[h]
	\centering
	\caption{Results of fine-grained classification with voting}
	\label{tab:fine-grained-voting}       
			\begin{tabular}{llccc}
				\textbf{Base Learner} & \textbf{Class} & $f_L$ & $f_L$+$f_S$ &  $f_L$+$f_S$+$f_M$ \\ \hline
				\multirow{9}{*}{NB, k-NB,RI,DT} & $F_{PER}$  & 79.81  & 81.67  & 82.86  \\
				& $F_{ORG}$  & 81.79  & 83.02  & 84.02  \\
				& $F_{LOC}$  & 81.97  & 83.74  & 84.91  \\
				& $F_{TEM}$  & 81.45  & 82.81  & 83.73  \\
				& $F_{NUM}$  & 81.83  & 82.07  & 83.54  \\
				& $F_{METH}$  & 82.15  & 83.13  & 84.09  \\
				& $F_{REA}$  & 82.24  & 83.36  & 84.42  \\
				& $F_{DEF}$  & 81.76  & 83.05  & 84.23  \\
				& $F_{MISC}$  & 80.21  & 82.33  & 83.21  \\ \hline
				
			\end{tabular}
\end{table}

\subsection{Result Analysis of Fine-Grained Classification} 

As research studies \cite{breiman-bagging:1996,clemen:1989,hansen-salamon:1990,hashem:1997} argued that classifier combination approaches provide better prediction results over individual classifier approach, our motivation is to verify the impact of the classifier combination approaches on Bengali QC task. 

Initially, we carried out our experiment with individual classifier approach and applied NB, k-NB, RI and DT classifiers separately.  Table~\ref{tab:fine-results-individual} presents the results obtained using individual classifier approach. In fine-grained classification task, we used the identical features that were also used in coarse-grained classification. Inevitably, the obtained accuracies for fine-grained classification is less than the coarse-grained classification using the same feature sets.

Then, we applied the state-of-the-art classifier combination techniques on the lexical, syntactic and semantic feature sets. Figure~\ref{fig:1} depicts the bagging size (i.e., number of iterations) for fine-grained classification. Breiman \cite{breiman-bagging:1996} stated that bagging approach improves the performance of a prediction model by reducing the variance. However, after certain iteration, it cannot reduce variance and the model becomes stable. Hence, after certain iteration, we were not able to improve the performance of the models. We noticed that the bagging approach requires more iteration to stable in fine-grained classification in comparison to the coarse-grained classification.  In contrast, boosting approach enhance the performance of the model by primarily reducing the bias \cite{schapire-boosting:1990}. After certain iterations, the boosting approach cannot reduce the bias because the corresponding weight of $\frac{1}{{\beta}_t}$ becomes less than 1. If $\frac{1}{{\beta}_t}$ is less than 1, then the weight of the classifier model in boosting may be less than zero for that iteration. Hence, in Figure~\ref{fig:2}, we can see that the boosting size is stable after certain iterations. Table~\ref{tab:fine-grained-ensemble} shows that the boosting approach achieves slightly better performance than the bagging.  In stacking approach, one classifier plays the role of ML and a set of classifiers act as BLs.  In the stacking approaches, the setup with NB, k-NB, RI as BLs and DT as ML outperforms other setup combinations. The stacking approach outperforms the voting approach with slight margin. However, the boosting approach with the base classifier DT achieves the best. It was noticed from the fine-grained question classification that all the classifier combination approaches beat the individual classifier approaches with a notable margin.

\subsection{Error Analysis}
We checked the dataset and the system output to analyze the errors.
We observed the following as the major sources of errors in the proposed system.
\begin{itemize}
    \item Questions belonging to different question classes have the same content words which make the classifiers confuse and wrongly classify the questions into same class. For example, both the questions ``koVna saByawAkeV bExika saByawA balA hayZa ?'' (gloss: which civilization is called the Vedic Civilization?), ``Arya saBya-wAkeV keVna bExika saByawA balA hayZa ?'' (gloss: why the Arya Civilization is called the Vedic Civilization?) have the same content words: \textit{saByawAkeV, bExika, saByawA, hayZa}.
    \item  In Bengli, the dual interrogatives consist of two single interrogatives. Thus, classifiers get confused by encountering two interrogative words. Therefore, classifiers often misclassify such questions.
    \item The classifiers wrongly classified the Bengali questions which are long and complex. For example, `keVna AXunika yugeVra paNdiweVrA maneV kareVna yeV, sinXu saByawA xrAbidZa jAwIra xbArA sqRti hayZeVCila ? (gloss: why the modern scholars think that the Indus Valley Civilization is created by the Aryans?).
\end{itemize}

\section{Conclusions}
\label{sec:conclusion}

Although QA research in other languages (such as English) has progressed significantly, for majority of Indian languages it is at the early stage of development. 
In this study, we addressed the QC task for Bengali, one of the most spoken languages in the world and the second most spoken language in India. 
We reported experiments for coarse-grained and fine-grained question classification. 
We employed lexical, syntactic and semantic features. 
We applied classifiers individually as well as combination approaches. 
The automated Bengali question classification system obtains up to 91.65\% accuracy for coarse-grained classes and 87.79\% for fine-grained classes using classifier combination approaches based on four classifiers, namely NB, k-NB, RI and DT. The contributions of this work are listed below.
\begin{itemize}
	\item This work successfully deploys state-of-the-art classifier combination approaches for the question classification task in Bengali.
    \item We have empirically established the efficacy of the classifier combination approach over individual classifier approach for coarse-grained question classification as well as fine-grained question classification.
    
    \item We have extended the single layered (coarse-grained) taxonomy  into two layered (coarse-grained and fine-grained) taxonomy by incorporating 69 fine-grained classes to the question classification taxonomy. 
	
	
    \item This work improves QC accuracy which in turns enhances the Bengali QA system performance.  
\end{itemize}
In coarse-grained question classification, overall the voting approach with majority voting technique performs best among the four classifier combination approaches, namely bagging, boosting, stacking, and voting. 
However, the stacking approach produces the best results for fine-grained classification. 

The only available QA dataset for Bengali contains only 1,100 questions. 
In future, we would like to contribute to enlarge the dataset.
One of the future directions of this study is employing the state-of-the-art neural network techniques. 
Also, we would like to apply the approaches used in this study to other less investigated languages.

\section{Acknowledgments}
Somnath Banerjee and Sudip Kumar Naskar are supported by Digital India Corporation (formerly Media Lab Asia), MeitY, Government of India, under the Visvesvaraya PhD Scheme for Electronics \& IT.

The work of Paolo Rosso was partially funded by the Spanish MICINN under the research
project PGC2018-096212-B-C31.


\end{document}